\documentclass[a4paper]{article}

\usepackage{INTERSPEECH2022}
\usepackage{multirow}
\usepackage{makecell}
\usepackage{svg}

\title{LAE: Language-Aware Encoder for Monolingual and Multilingual ASR}
\name{Jinchuan Tian$^{1,2}$\thanks{This work is done when Jinchuan Tian was an intern in Tencent AI Lab.},
Jianwei Yu$^{2,3}$, 
Chunlei Zhang$^{2}$, 
Chao Weng$^{2,3}$, 
Yuexian Zou$^{1}$, 
Dong Yu$^{2}$
}

\address{$^1$ADSPLAB, School of ECE, Peking University, Shenzhen, China \\
         $^2$Tencent AI Lab, $^3$Tencent ASR Oteam}
\email{tomasyu@tencent.com; zouyx@pku.edu.cn}

\begin{document}

\maketitle
\begin{abstract}
Despite the rapid progress in automatic speech recognition (ASR) research, recognizing multilingual speech using a unified ASR system remains highly challenging.
Previous works on multilingual speech recognition mainly focus on two directions: recognizing multiple monolingual speech or recognizing code-switched speech that uses different languages interchangeably within a single utterance.  
However, a pragmatic multilingual recognizer is expected to be compatible with both directions. 
In this work, a novel language-aware encoder (LAE) architecture is proposed to handle both situations by disentangling language-specific information and generating frame-level language-aware representations during encoding.
In the LAE, the primary encoding is implemented by the shared block while the language-specific blocks are used to extract specific representations for each language.
To learn language-specific information discriminatively, a language-aware training method is proposed to optimize the language-specific blocks in LAE.
Experiments conducted on Mandarin-English code-switched speech suggest that the proposed LAE is capable of discriminating different languages in frame-level and shows superior performance on both monolingual and multilingual ASR tasks. 
With either a real-recorded or simulated code-switched dataset, the proposed LAE achieves statistically significant improvements on both CTC and neural transducer systems.
Code is released\footnote{https://github.com/jctian98/e2e\_lfmmi/egs/asrucs}.
\end{abstract}
\vspace{-5pt}

\section{Introduction}
As more than 60\% of the population in the world are multilinguists\cite{population}, the automatic speech recognition (ASR) system for multilingual speech is in urgent need.
Although rapid progress has been made in monolingual speech recognition, the recognition of multilingual speech still remains a challenging task\cite{survey}.
In real applications, a pragmatic recognizer is commonly expected to address two scenarios simultaneously.
One is the monolingual speech from multiple languages, in which only one language is used within each utterance but the languages used in different utterances vary. The other is the code-switched speech, in which multiple languages are used interchangeably within a single utterance.

Most of the previous works mainly focus on monolingual speech from various languages or the code-switched speech one or the other. For the former situation, the monolingual speech from multiple languages is handled by mixing multiple monolingual corpora during training\cite{mix_train, mix_train2}. 
The latter situation is mainly addressed by approaches like predicting language-switching points\cite{switch2} or identifying language-IDs\cite{langid, lid1}, building unified acoustic\cite{ssl1, ssl2} or textual\cite{bytes} representations for all considered languages, 
integrating language models customized for code-switch ASR\cite{mono3, lm1, lm2}, incorporating external monolingual speech and text resources\cite{mono1, mono2, mono3}, or enriching the multilingual corpora by simulated training utterances\cite{simu1, simu2, simu3, simu4}. 
To our knowledge, the works that indiscriminately consider both scenarios are still limited\cite{population, moe, 21, brian}. 
Our previous work \cite{brian} addresses this problem by factorizing the multilingual ASR task into monolingual sub-tasks and fine-tuning on encoders that are inter-independently pre-trained over the monolingual corpora. Although considerable improvement is achieved in \cite{brian}, we argue the pre-training process is time-consuming and its performance is compromised due to the lack of interaction among the separated encoders.

To recognize both monolingual and code-switched speech consistently, a desired property of the system is to disentangle the language-specific information and generate frame-level language-aware representations during the encoding stage. 
Then these representations can be partially used in the monolingual situations and fully explored during the recognition of code-switched speech. 
In this work, a novel language-aware encoder (LAE) architecture is proposed, in which the utterance is firstly encoded by the shared block and the language-specific representations for each language are then extracted by several language-specific blocks. 
Additionally, a new language-aware training method enables the model to discriminate different language-specific information in both utterance-level and frame-level.
For the recognition of each mono-language utterance, only one of the language-specific blocks is informative while the others stay implicitly idle. 
For the code-switched speech, the whole LAE is activated and the utterance-level encoder output is interlaced from the frame-level language-aware representations of all languages.
The proposed LAE is applicable to two of the widely used ASR frameworks: connectionist temporal classification (CTC) \cite{ctc} and neural transducers (NTs) \cite{rnnt}. 
Experiments conducted on ASRU 2019 Mandarin-English Code-Switch Challenge dataset suggest that the adoption of LAE architecture outperforms our previous work \cite{brian} and achieves statistically significant improvements: for models trained with either real-recorded or simulated code-switched data, up to 23.3\% and 18.8\% mixed error rate (MER) reductions are obtained.

Our main contributions are summarized as follows:
Firstly, we propose a novel language-aware encoder (LAE) architecture, which is compatible with both multiple monolingual and code-switched speech. In contrast, most of the previous works mainly focus on one of these two scenarios.
Secondly, a language-aware training method is provided and the LAE is allowed to discriminate language-specific information in both utterance-level and frame-level.
Such methods can significantly boost the performance of multilingual ASR systems with or without real-recorded code-switched data.

\section{Language-Aware Encoder}
In this section, the proposed language-aware encoder (LAE) architecture, the language-aware training method and the unified decoding workflow are introduced in the following three parts respectively.
\vspace{-3pt}
\subsection{Language-Aware Encoder Architecture}
\vspace{-3pt}
\label{sec_architecture}
Fig.1 demonstrates a Mandarin-English bilingual speech recognizer with the proposed LAE architecture.
The input frames are firstly fed into the shared block followed by the parallel Mandarin and English blocks. The hidden representations from the two language-specific blocks are then combined as the LAE hidden outputs $\mathbf{h}_{\text{bil}}$ and then sent to the global decoder to facilitate speech recognition.
The shared encoder is adopted to encode the input frames globally and provide more interaction in primary stage of encoding.
In each of the language-specific blocks, information from the target language is captured while the non-target information is expected to be filtered out.

\vspace{-3pt}
\subsection{Language-Aware Training}
\vspace{-3pt}
\label{sec_mono_super}
Although the language-specific blocks are designed with the goal of disentangling language-specific representations, there is no explicit guarantee for this discriminative property as the encoding process of these blocks is symmetric.
To allow the language-specific blocks to discriminatively extract the information from the target language and filter out the undesired information from other languages, the language-aware training method is proposed.

During training, a monolingual Mandarin token sequence $Y_{\text{man}}$ is generated from the bilingual target token sequence $Y$ by masking every English token with a special token \textit{\textless eng\textgreater}. 
Similarly, a monolingual English token sequence $Y_{\text{eng}}$ is generated with the special token \textit{\textless man\textgreater}.
Next, the monolingual Mandarin token sequence is considered as the target to optimize the output of Mandarin-specific decoder, which is an auxiliary criterion during training:
\begin{equation}
\setlength\abovedisplayskip{0.2cm}
\setlength\belowdisplayskip{0.2cm}
     J_{\text{man}} = \text{Aux\_Criterion}(Y_{\text{man}}, \text{Decoder}_{\text{man}}(\mathbf{h}_{\text{man}}))
\end{equation}
A similar auxiliary criterion is also used in the English side:
\begin{equation}
\setlength\abovedisplayskip{0.2cm}
\setlength\belowdisplayskip{0.2cm}
     J_{\text{eng}} = \text{Aux\_Criterion}(Y_{\text{eng}}, \text{Decoder}_{\text{eng}}(\mathbf{h}_{\text{eng}}))
\end{equation}
Finally, the global training objective is the interpolation of the original training criterion $J_{\text{ori}}$ and the mean of these auxiliary criteria.
\begin{equation}
\setlength\abovedisplayskip{0.2cm}
\setlength\belowdisplayskip{0.2cm}
     J = J_{\text{ori}} + (J_{\text{man}} + J_{\text{eng}}) / 2
\end{equation}
where the original training criterion $J_{\text{ori}}$ can be either CTC criterion or transducer criterion according to the choice of the global decoder architecture.

In this work, the CTC criterion is consistently adopted as the auxiliary criterion for simplicity and the language-specific decoders are linear classifiers. 
We share the parameters of these language-specific decoders. As there is no overlap between the sequences $Y_{\text{man}}$ and $Y_{\text{eng}}$, sharing these parameters will not result in a leak of supervision information and can achieve better alignment between the Mandarin and English blocks in the prediction of CTC \textit{\textless blank\textgreater}. Also, we make $Y$, $Y_{\text{man}}$ and $Y_{\text{eng}}$ in the same lengths in the hope that the language-specific blocks are able to predict the number of tokens when being inactive.

\vspace{-3pt}
\subsection{Unified decoding with global decoder}
\vspace{-3pt}
As shown in Fig.2, for code-switched input, the proposed method is capable to disentangle the Mandarin and English segments in both monolingual and code-switched speech.  
Frames belonging to English and Mandarin can be extracted in English block outputs $\mathbf{h}_{\text{man}}$ and Mandarin block outputs $\mathbf{h}_{\text{eng}}$ respectively.
For the monolingual input, the corresponding language-specific block is activated to predict ordinary token sequence while the other language-specific blocks are implicitly idle: they only predict the number of tokens by predicting repetitive masking symbols.
Since there will be limited overlapping between the Mandarin and English frames, it is reasonable to simply combine $\mathbf{h}_{\text{man}}$ and $\mathbf{h}_{\text{eng}}$ with the frame-level addition:
\begin{equation}
    \setlength\abovedisplayskip{0cm}
    \setlength\belowdisplayskip{0cm}
    \mathbf{h}_{\text{bil}} = \mathbf{h}_{\text{man}} \oplus \mathbf{h}_{\text{eng}}
\end{equation}
$\mathbf{h}_{\text{bil}}$ keeps the original order of the English and Mandarin segments and keeps the highly representative nature of  $\mathbf{h}_{\text{man}}$ and  $\mathbf{h}_{\text{eng}}$ for decoding.
With $\mathbf{h}_{\text{bil}}$, the global decoder can be used to recognize both the Mandarin and English monolingual speech and the code-switch speech in a unified manner. 

\begin{figure}
    \centering
    \setlength\abovedisplayskip{0cm}
    \setlength\belowdisplayskip{0cm}
    \includegraphics[width=\linewidth]{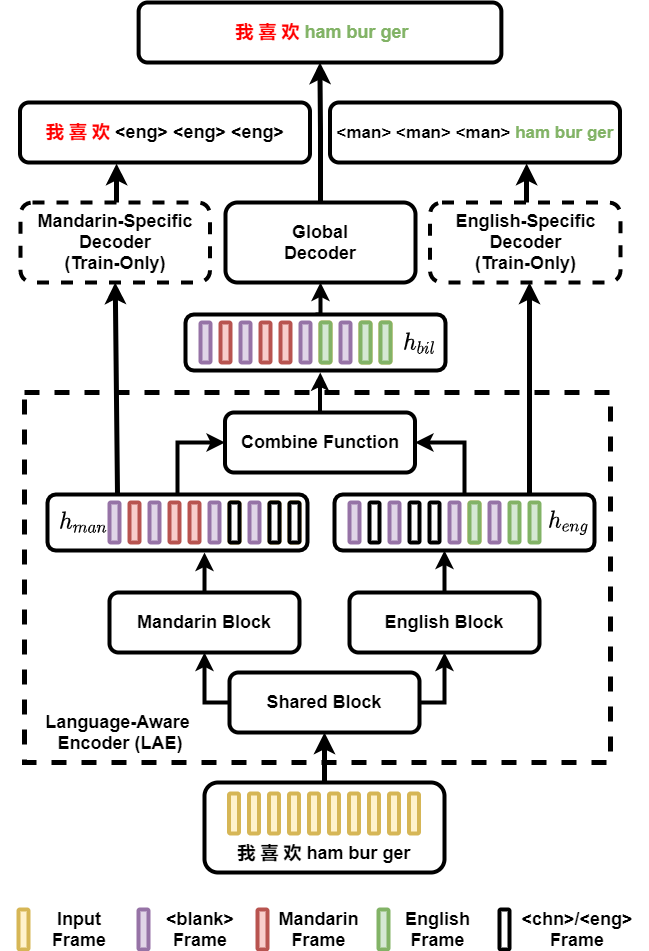}
    \caption{Mandarin-English Bilingual Speech Recognizer with Proposed Language-Aware Encoder (LAE) Architecture. (English translation of input utterance: \textit{I like hamburger})}
    \label{fig_lae}
    \vspace{-20pt}
\end{figure}

\begin{figure*}[t]
\centering
\begin{minipage}[t]{0.33\textwidth}
\centering
\includegraphics[width=1\textwidth]{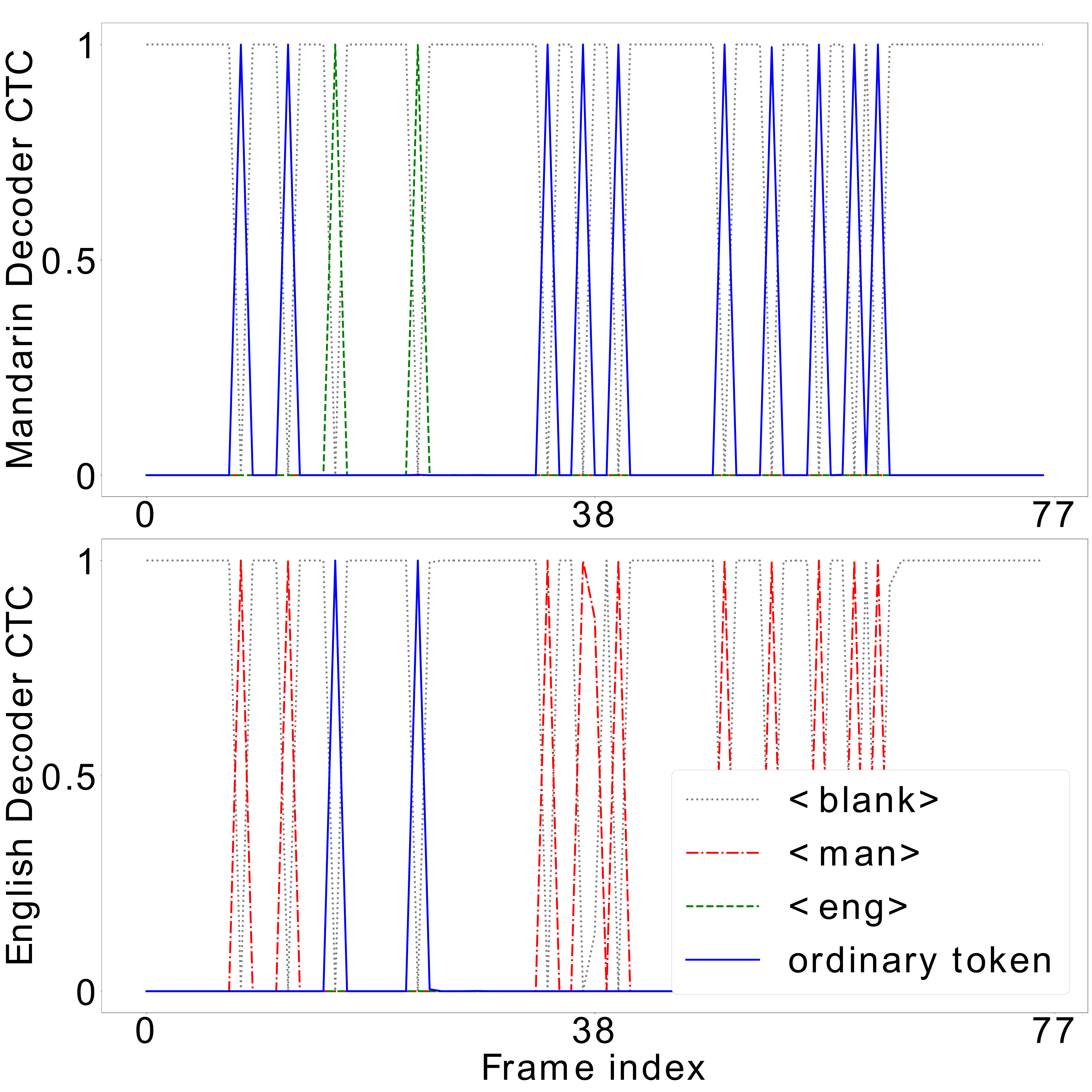}
\end{minipage}
\begin{minipage}[t]{0.33\textwidth}
\centering
\includegraphics[width=1\textwidth]{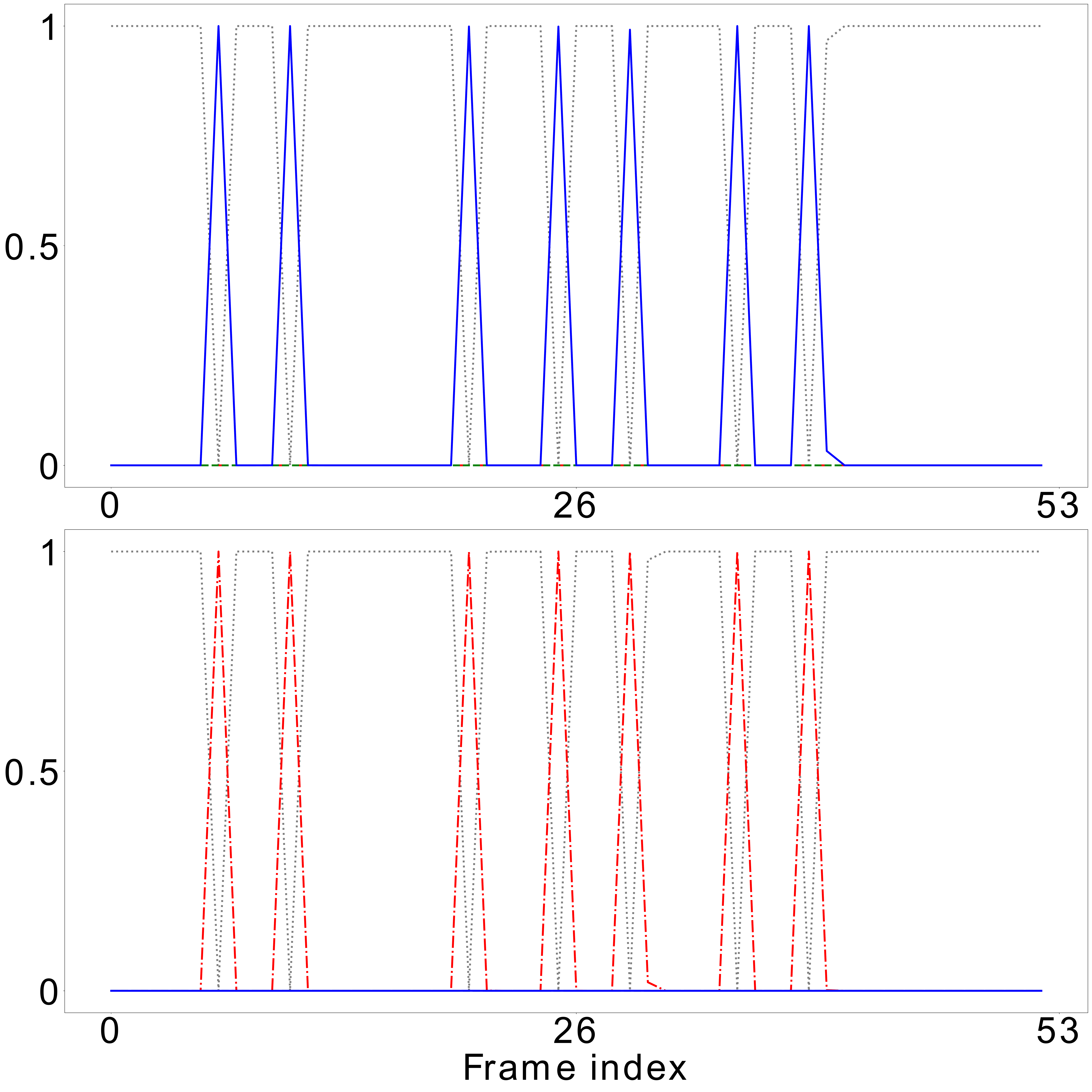}
\end{minipage}
\begin{minipage}[t]{0.33\textwidth}
\centering
\includegraphics[width=1\textwidth]{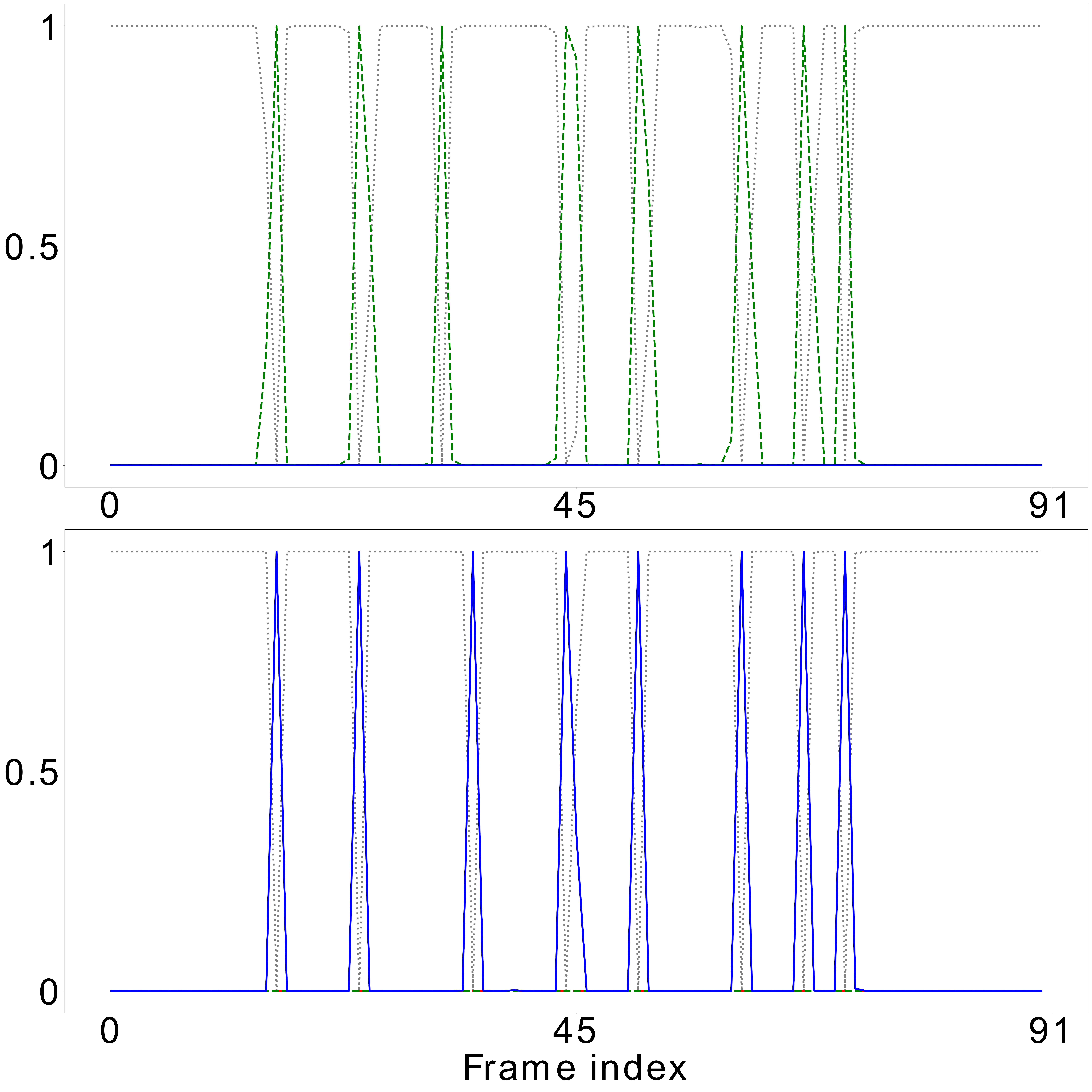}
\end{minipage}
\caption{CTC distributions generated by CTC Mandarin and English-specific decoders for different input speech. Left: code-switch input; middle: monolingual Mandarin input; right: monolingual English input.}
\label{fig_ctc}
\vspace{-20pt}
\end{figure*}

\section{Experiments}

In this section, the experimental setup is introduced in section \ref{exp_setup}.
The performance of the proposed method is firstly verified on the real-recorded code-switch speech dataset in section \ref{exp_real}. The model's ability to discriminate languages is demonstrated in section \ref{exp_abla}. As the scarcity of multilingual data is widely concerned and data simulation methods are generally adopted for alleviation, it is of great importance to confirm that the proposed method can benefit from the simulated multilingual data. 
To this end, experiments on the simulated code-switch speech are further conducted in section \ref{exp_simu}.
\vspace{-5pt}
\subsection{Experimental Setup}
\vspace{-5pt}
\label{exp_setup}
All experiments are conducted on the ASRU 2019 Mandarin-English Code-Switch dataset\cite{data}, in which 200h code-switched speech and 500h Mandarin speech are provided. Following \cite{brian}, 500h accented English data\footnote{http://speechocean.com} is also used as the mono-English data. By default, all monolingual utterances are used in training.
A simulated code-switch speech training set called \textit{Simu-CS} is additionally generated by simply splicing the raw waves and the corresponding transcriptions from the monolingual speech. This simulated data also has the size of 200h and the length of each utterance is below 12 seconds.
The code-switch evaluation set is officially provided and the monolingual evaluation sets are generated following \cite{brian}.
For both global decoder and language-specific decoders, the predictions are based on a fixed vocabulary, which contains 4294 Mandarin characters, 5000 English BPE \cite{bpe} tokens and three special tokens: \textit{\textless blank\textgreater}, \textit{\textless man\textgreater} and \textit{\textless eng\textgreater}.

The input features are 80-dim Fbank features plus 3-dim pitch features. These features are sub-sampled with a factor of 4 by a two-layer CNN before being fed into the encoders. 
All encoders are stacked Conformer\cite{conformer} layers, in which the attention dimension, feed-forward dimension, number of attention heads, number of convolutional kernels are fixed to 512, 2048, 4, 31 respectively.
Three encoder architectures with similar parameter budgets are designed for comparison:
\textbf{Vallina}: 15 stacked Conformer layers.
\textbf{Bi-Encoder}: two separated encoders with 8 Conformer layers stacked each.
\textbf{Language-Aware Encoder (LAE)}: the shared block contains 9 conformer layers while the language-specific blocks consist of 3 Conformer layers each.
For the CTC system, the encoder output is linearly classified to make predictions. 
The global decoder of the NT system consists of a 2048-dim LSTM prediction network and a 256-dim MLP joint network.

All models are optimized by Adam optimizer with warm-up steps of 25k,  peak learning rate of 3e-4 and inverse square root decay schedule. 
The training process lasts for 100 epochs and is implemented on 8 Tesla V100 GPUs. 
The effective batch size is 512 with a gradient accumulation number of 8. 
SpecAugment\cite{specaug} is consistently used with 2 time masks and 2 frequency masks.
The checkpoints from the last 10 epochs are averaged before evaluation. 
Standard CTC prefix search and ALSD algorithm \cite{alsd} are adopted for CTC and NT decoding respectively with a fixed beam size of 10.
A word-level 5-gram language model\cite{wngram} trained from all training transcriptions is optionally used with a fixed weight of 0.2. 
We report both code-switched and monolingual results following \cite{brian}.
Code is mainly revised from Espnet\cite{espnet}.

\begin{table*}[]
    \centering
    \scalebox{0.8}{
    \begin{tabular}{c|l|c|c|c|c|c}
    \toprule
         \multirow{2}{*}{No.} & \multirow{2}{*}{System} & \multicolumn{3}{|c|}{Code-Switch} & \multicolumn{2}{|c}{Monolingual} \\
    \cline{3-7}
    && All (MER) & Man (CER) & Eng (WER) & Man (CER) & Eng (WER) \\
    \hline
    \multicolumn{7}{l}{Literature} \\
    \hline
    & Single NT \cite{21, population}                               & 11.3 & 9.3 & 30.8 & 6.5 & 17.8 \\
    & Gating NT \cite{moe, population}                               & 11.2 & 8.8 & 34.7 & 5.7 & 34.6 \\
    & Factorized NT + Pre-train\cite{brian}                & 10.2 & 8.1 & 29.2 & 5.3 & 16.3 \\
    & Hybrid CTC-Attention \cite{shuaizhang}               & 9.5  & 7.6 & 25.0 & -   & -    \\
    \hline
    \multicolumn{7}{l}{Our Results} \\
    \hline
    1 & Vallina CTC (baseline)                             & 11.6 & 9.2 & 38.7 & 5.1 & 20.3 \\
    2 & Bi-Encoder CTC                                     & 10.3 & 8.1 & 36.1 & 3.6 & 19.4 \\ 
    3 & \ \ + Language-Aware Training                & 10.2 & 8.1 & 34.7 & 3.6 & 19.1 \\
    4 & LAE CTC                                            & 10.3 & 8.2 & 34.9 & 3.5 & 18.5 \\
    5 & \ \ + Language-Aware Training (proposed)     & 9.5  & 7.5 & 33.1 & 3.2 & 17.7 \\
    6 & \ \ \ \ + word N-gram LM                           & \bf{8.9*}  & \bf{7.0*} & \bf{30.9*} & \bf{2.4*} & \bf{17.0*} \\
    \hline
    7 & Vallina NT (baseline)                              & 9.5 & 7.9 & 28.6 & 3.8 & 16.6\\
    8 & Bi-Encoder NT                                      & 9.9 & 8.2 & 29.3 & 3.9 & 17.7 \\
    9 & LAE NT + Language-Aware Training (proposed)          & 9.1 & 7.5 & 28.2 & 3.4 & 16.4 \\
    10&\ \ + N-gram LM                                & \bf{8.9*} & \bf{7.3*} & \bf{27.7*} & \bf{2.7*} & \bf{15.9*} \\
    \bottomrule           
    \end{tabular}
    }
    \caption{Performance of models trained on both monolingual and real-recorded code-switch data. * means statistically significant improvement compared with the baseline systems.}
    \label{tab_real}
    \vspace{-20pt}
\end{table*}

\vspace{-5pt}
\subsection{Results with real-recorded code-switch speech}
\vspace{-5pt}
\label{exp_real}

The results with real-recorded code-switch and monolingual training data are reported in table \ref{tab_real}. Our main observations are reported as follows.

\noindent\textbf{Overall performance}: 
As shown in table \ref{tab_real}, the proposed method achieves significant performance improvement on both CTC and NT systems. 
For the CTC system, the relative improvements on code-switch, mono-Mandarin and mono-English evaluation sets are 23.3\%, 52.9\% and 16.3\% respectively (exp.1 vs. exp.6) while these numbers for the NT system are 6.3\%, 28.9\% and 4.2\% (exp.7 vs. exp.10).
The significance of these improvements is further confirmed by matched pairs sentence-segment word error (MAPSSWE) based significant test\footnote{https://github.com/talhanai/wer-sigtest} with the significant level of $p=0.001$.
The results in exp.6 and exp.10 verify that the proposed method is applicable to both monolingual and code-switch scenarios simultaneously. 

\noindent\textbf{Language-Aware Training for LAE}:
Compared with the vallina CTC system, considerable improvement is provided by the LAE architecture (exp.1 vs. exp.4). Similar improvement is also achieved by the Bi-Encoder architecture (exp.1 vs. exp.2), which emphasizes the necessity of handling each language by a specific block. 
Further study shows that the error reduction provided by language-aware training in Bi-Encoder architecture is marginal (exp.2 vs. exp.3, also observed in \cite{brian}) while the adoption of language-aware training on the proposed LAE system provides up to 0.8\% absolute improvement (exp.4 vs. exp.5). This observation implies the adoption of the shared block is beneficial.
The adoption of LAE architecture and language-aware training also provides up to 1.3\% absolute improvement on NT systems (exp.8 vs. exp.9). 
In summary, the combination of the LAE and the language-aware training shows superior performance on both CTC and NT systems.

\noindent\textbf{Comparison between CTC and NT system}. 
With the same LAE architecture, the NT system outperforms the CTC system if no language models are integrated (exp.5 vs. exp.9). 
As the performance differences are mainly in the English scenarios, this phenomenon can be attributed to the well-known conditional independence assumption of the CTC system\cite{ctc}: English words need more context dependency to spell words correctly. 
Further investigation demonstrates that the language models are more beneficial on the CTC system than on the NT system (exp.6 vs. exp.10), so we suppose the conditional-independence of CTC is alleviated by the context information in language models.   
\vspace{-5pt}
\subsection{Ability to discriminate languages}
\vspace{-5pt}
\label{exp_abla}
In this part, we experimentally show that the proposed model is capable to discriminate the language-specific representations in both utterance-level and frame-level.

For utterance-level capability, an additional language classifier is applied to the time-dimensional mean value of LAE hidden output $\mathbf{h}_{\text{bil}}$ during training to tell an utterance is mono-English, mono-Mandarin or code-switched. As suggested in table \ref{tab_lag_cls}, the accuracy over the three evaluation sets is all above 99.6\%, which implies the output of LAE provides strong evidence to classify the utterance.

\begin{table}[]
    \centering
    \begin{tabular}{p{2cm}<{\centering}|p{2cm}<{\centering}|p{2cm}<{\centering}}
    \toprule
    \multicolumn{3}{c}{Language Classification (Acc\%)} \\ 
    \hline
    Code-Switch & Mandarin & English \\
    \hline
    99.65     & 99.91 & 99.98 \\
    \bottomrule
    \end{tabular}
    \caption{Utterance-level language classification accuracy of the proposed model over the three evaluation sets}
    \label{tab_lag_cls}
    \vspace{-20pt}
\end{table}

\begin{table}[]
    \centering
    \scalebox{0.9}{
    \begin{tabular}{|p{1.5cm}<{\centering}|p{1.5cm}<{\centering}|p{1.5cm}<{\centering}|p{1.5cm}<{\centering}|}
    \hline
         \multicolumn{2}{|c|}{Subset of Code-Switch} & \multicolumn{2}{|c|}{Monolingual} \\
         \hline
         Man (CER) & Eng (WER) & Man (CER) & Eng (WER) \\
         \hline
         \multicolumn{4}{|l|}{Global Decoder (Reference)} \\ 
         \hline
         7.5 & 33.1 & 3.2 & 17.7 \\
         \hline
         \multicolumn{4}{|l|}{Mandarin-Specific Decoder} \\ 
         \hline
         9.3 & 100.0 & 3.5 & 100.0 \\
         \hline
         \multicolumn{4}{|l|}{English-Specific Decoder} \\ 
         \hline
         98.3 & 35.0 & 99.7 & 17.5 \\
         \hline
         
    \end{tabular}
    }
    \caption{Results of decoding using the language-specific decoders and the global decoder}
    \vspace{-8pt}
    \label{tab_axu_decode}
    \vspace{-20pt}
\end{table}

To demonstrate that the LAE is also capable to discriminate language-specific information in frame-level, the decoding results of language-specific decoders are compared with those results of the global decoder in table \ref{tab_axu_decode}. 
For both code-switch and monolingual situations, the language-specific encoders achieve comparable results with those of global encoders in the corresponding target languages but fail in other language. This observation proves that the hidden output $\mathbf{h}_{\text{man}}$ and $\mathbf{h}_{\text{eng}}$ are discriminative in frame-level: they keep sufficient information from the target language for independent decoding while filtering out the information from non-target languages. 

Visualization is provided in Fig.\ref{fig_ctc} to further exemplify the language-specific representations in frame-level. 
Firstly, the language-specific encoders are well-aligned in time-dimensional and the CTC make spikes simultaneously. This property is desired to alleviate the \textit{token-or-blank} conflicts in the feature combination stage.
Secondly, whenever a Mandarin token is predicted by the Mandarin-specific decoder, the mask token is predicted by the English-specific decoder and vice versa. This verifies that the language-specific blocks can cooperate consistently: they only capture the information from the specific language and are activated interchangeably when the language-switching happens.
\vspace{-5pt}
\subsection{Results with simulated multilingual speech}
\vspace{-3pt}
\label{exp_simu}
In this part, we further demonstrate the proposed LAE method can benefit from the simulated code-switch data. Experimental results are presented in table \ref{tab_simu}.
\vspace{-10pt}
\begin{table}[htpb]
    \centering
    \scalebox{0.8}{
    \begin{tabular}{|l|c|c|c|c|c|}
    \hline
         \multirow{2}{*}{System} & \multicolumn{3}{|c|}{Code-Switch} & \multicolumn{2}{|c|}{Monolingual} \\
    \cline{2-6}
    & All & Man & Eng & Man & Eng \\
    \hline
    Vallina CTC                   & 34.0 & 35.5 & 102.5 & 5.8 & 20.6\\
    \ \ + Simu-CS                 & 38.7 & 39.4 & 114.8 & 6.6 & 20.4\\
    LAE CTC + Lang.-Aware Training  & 32.6 & 33.4 & 93.3 & 4.1 & 18.8 \\
    \ \ + Simu-CS                 & 29.5 & 29.3 & 89.1 & 4.0 & 18.5 \\
    \ \ \ \ + N-gram LM           & \bf{27.6} & \bf{27.8} & \bf{87.5} & \bf{3.1} & \bf{17.6} \\
    \hline
    \end{tabular}
    }
    \caption{Performances of models trained from monolingual data. The simulated code-switch data is optionally used.}
    \label{tab_simu}
    \vspace{-20pt}
\end{table}

For the LAE CTC system, the adoption of simulated code-switch training data achieves an absolute error reduction of 3.1\% over the code-switch evaluation set. Further experiments with language models reduce the MER from 34.0\% to 27.6\%, which is a 18.8\% relative improvement. By contrast, no performance improvement is observed on the vallina CTC system when the simulated data is added. We suppose the simply stacked Conformer layers can hardly capture the dynamics of language-switching in the simulated data.
\vspace{-5pt}
\section{Conclusion}
\vspace{-3pt}
The language-aware encoder (LAE) and the language-aware training are proposed in this work to address monolingual and multilingual speech problems simultaneously.
The proposed model is capable to capture the language-specific features of each language by specific encoder blocks. 
The proposed method is verified compatible with both CTC and neural transducer systems. 
Experimentally, our models achieve statistically significant improvements in both monolingual and multilingual scenarios and show their strengths on simulated bilingual corpora.

\bibliographystyle{IEEEtran}
\bibliography{./mybib.bib}

\end{document}